\documentclass[runningheads]{llncs}
\usepackage{graphicx}
\begin{document}
\title{Aligning Visual and Lexical Semantics}
\titlerunning{Aligning Visual and Lexical Semantics}

\author{Fausto Giunchiglia \orcidID{0000-0002-5903-6150} \and
Mayukh Bagchi \orcidID{0000-0002-2946-5018} \and
Xiaolei Diao \orcidID{0000-0002-3269-8103}}
\authorrunning{Fausto Giunchiglia et al.}
\institute{Department of Information Engineering and Computer Science (DISI) \\ University of Trento, Trento, Italy.
\email{\{fausto.giunchiglia,mayukh.bagchi,xiaolei.diao\}@unitn.it}\\}
\maketitle              
\begin{abstract}
We discuss two kinds of semantics relevant to Computer Vision (CV) systems - \emph{Visual Semantics} and \emph{Lexical Semantics}. While visual semantics focus on how humans build concepts when using vision to perceive a target reality, lexical semantics focus on how humans build concepts of the same target reality through the use of language. The lack of coincidence between visual and lexical semantics, in turn, has a major impact on CV systems in the form of the Semantic Gap Problem (SGP). The paper, while extensively exemplifying the lack of coincidence as above, introduces a general, domain-agnostic methodology to enforce alignment between visual and lexical semantics.

\end{abstract}

\section{Introduction}
\label{S1}
Let us begin with a motivating example from the domain of musical instruments. The musical instrument \emph{Koto}, for instance, can be visually perceived in potentially an infinite number of ways. It can be, for example, perceived as a generic \emph{musical instrument} or as a \emph{stringed musical instrument} by someone who is not familiar with Japanese music. It can also be (mis)perceived as a Chinese \emph{Zheng} or a Mongolian \emph{Yatga} or a Kazakhstani \emph{Jetigen} by individuals closer to Chinese, Mongolian or Kazakhstani culture. Those who are experts in Japanese music can perceive it as a \emph{Gakuso, Chikuso, Zokuso} or \emph{Tagenso} (types of \emph{Koto}). Finally, the \emph{Koto} can also be \emph{Koto\#2010}, a specific \emph{Koto} which my music academy bought in 2010. However, the interesting observation is the fact that each of the above (visual) perceptions can subsequently be encoded linguistically in, again, potentially an infinite number of ways. Thus, for example, a \emph{Koto} can be linguistically encoded as a \emph{Koto} or as a \emph{Kin} or as a \emph{Jusangen} (all of them being synonyms) in Japanese \cite{KOTO}. The same \emph{Koto}, however, being a \emph{lexical gap} \cite{UKC-CICLING} in any of the Indian languages can only be defined as a board zither with 13 strings and movable bridges, without any assignment of a linguistic \emph{label}.

We observe from the above motivating example that given any target reality, there is a clear and \emph{unavoidable} lack of alignment between how humans visually perceive the target reality and how they linguistically interpret it for communication and reasoning \cite{2017-ER,SNCS-2021}. In fact, the aforementioned lack of alignment is pervasive in Computer Vision (CV) systems and has been crystallized as the \emph{Semantic Gap Problem} (SGP) \cite{SGP}, \emph{viz.} the lack of coincidence between a visual data (such as object(s) in an \emph{image}) and its user linguistic interpretation in a specific context. Thus, for instance, given the perceived visual data as an image of a \emph{Koto}, it is not necessarily the case that its linguistic interpretation is codified as exactly that of a \emph{Koto} in a particular context. 

We recast the SGP to be essentially the problem of a many-to-many mapping between \emph{Visual Semantics} (encoding the visual data as in SGP) and \emph{Lexical Semantics} (encoding the user linguistic interpretation as in SGP) which generates the lack of alignment \cite{2021-CAOS}. While visual semantics concentrate on how humans recognize and mentally internalize concepts via visually perceiving a target reality, lexical semantics is instead focused on how concepts are constructed by humans via linguistically encoding mental representations of a target reality in the form of words and word meanings in a particular language \cite{UKC-IJCAI,UKC-CICLING}. For instance, given the target reality of a specific instance of \emph{Koto}, the visual semantics of a CV system (such as a \emph{smartglass}) can perceive and encode the \emph{Koto} at different levels of abstraction corresponding to different \emph{partial views}, each of which can further be linguistically codified in different ways in different languages. 

Towards resolving such misalignment, the goal of this paper is to \emph{introduce a general, domain-agnostic methodology which enforces the alignment between visual and lexical semantics}. The key \emph{novelty} of the methodology lies in the fact that, differently from mainstream CV approaches (see, for example, \cite{ONTOCV,ICML-2020}), the very basis of computing visual semantics (via \emph{visual properties}) \cite{SNCS-2021} is independent from that of computing lexical semantics (via \emph{linguistically grounded properties}), while still being functionally linked and aligned.

The remainder of the paper is organized as follows. Section \ref{S2} discusses and exemplifies the notions of visual semantics and lexical semantics. Section \ref{S3} details the SGP in terms of the misalignment of visual and lexical semantics. Section \ref{S4} illustrates the proposed methodology to align visual and lexical semantics and thus resolve the SGP. Section \ref{S5} concludes the paper.

\section{Visual \& Lexical Semantics}
\label{S2}
We assume that reality is composed of \emph{substances}, where substances are things \emph{``about which you can learn from one encounter something of what to expect on other encounters, where this is no accident but the result of a real connection"} \cite{millikan2000}. Grounded in the above assumption, visual semantics focus on how \emph{``humans build concepts when using vision"} \cite{SNCS-2021} to perceive a target reality of substances. Such concepts generated from visual experientiality \cite{2021-CAOS,ISKO}, are termed \emph{Substance Concepts} (see \cite{2016-FOIS}). Notice that substance concepts are always formed from \emph{partial views} of a substance, such partiality arising from factors such as different focus or different confounding variables such as occlusion, clutter etc. For example, let us continue with the motivating case of a \emph{Koto} which, in our terms, is a substance. Further, according to the theory above, the same substance \emph{Koto} can generate different substance concepts depending on the different partial views. For a member in the audience of a concert where the \emph{Koto} is played, the substance concept can be just a \emph{stringed musical instrument} because he/she recognizes that, differently from a keyboard, the \emph{Koto} is played via taut strings. For the concertmaster, the same instrument can generate the substance concept of a \emph{Koto} because he/she recognizes that a \emph{Koto} has thirteen taut strings. For the musician playing the \emph{Koto}, instead, the substance concept generated can be \emph{my koto} which he/she bought in early 2021. 

There are three characteristics of visual semantics, as exemplified above, which are of key significance to CV systems. Firstly, the fact that visual semantics is computed over the partially viewable \emph{visual properties} \cite{SNCS-2021} (e.g., thirteen taut strings) of a substance and not over linguistically grounded properties as is standard in mainstream CV benchmarks (see, for instance, \cite{IMAGENET-2009,coco}). Secondly, the computation of visual semantics is \emph{hierarchical} by nature, i.e., it \emph{implies} the continuous construction of a \emph{visual subsumption hierarchy} \cite{SNCS-2021} which taxonomically interrelates, over several levels, general and specific substance concepts. For example, the same \emph{Koto} can generate a visual subsumption hierarchy where the most general substance concept is \emph{stringed musical instrument}, the intermediate substance concept is \emph{Koto} and the most specific substance concept is \emph{Tagenso}. Thirdly and finally, the important observation that the visual subsumption hierarchy is constructed by exploiting \emph{visual genus-differentiae} \cite{SNCS-2021} over visual properties. For example, while a \emph{stringed musical instrument} is recognized via its visual genus of \emph{taut strings}, the visual differentia of \emph{number of taut strings} taxonomically differentiates a \emph{stringed musical instrument} into a \emph{guitar} (six taut strings) or a \emph{koto} (thirteen taut strings).

Lexical semantics, on the other hand, focus on how humans build concepts through the use of language (termed \emph{Classification Concepts} as from \cite{2016-FOIS}) for a particular target reality. It does so by constructing and exploiting large-scale (multilingual) lexical resources which are general (see, for instance, Princeton WordNet (PWN) \cite{PWN}) or domain-specific \cite{WND} in scope. The fundamental language constructs which model the classification concepts composing such resources include word, word sense (encoding a specific meaning of a word), \emph{synsets} (a set of synonyms having same word sense), \emph{glosses} (a natural language definition encoding the meaning of a synset) and examples (exemplifying a gloss). For example, in the musical instrument domain hierarchy of a lexical resource, the word \emph{Koto} can be the preferred term of a synset comprising of the synonyms - \emph{Koto}, \emph{Kin} and \emph{Jusangen}. The synset \emph{Koto} can be defined as \emph{``a Japanese stringed instrument, consisting of a rectangular wooden body over which are stretched silk strings, which are plucked with plectrums or a nail-like device"}\footnote{https://www.collinsdictionary.com/dictionary/english/koto}. 

In sync with visual semantics, there are three key characteristics of lexical semantics which are noteworthy. Firstly, the fact that lexical semantics is computed over \emph{linguistically grounded properties} of a substance which might not necessarily be visually perceptible. For example, the property of a \emph{Koto} being made of a specific kind of Paulownia wood is \emph{extravisual} in nature, the details of which can only be defined linguistically. Secondly, \emph{lexical subsumption hierarchies} \cite{UKC-IJCAI} are computed based on linguistically grounded properties, which interrelates, over levels, generic classification concepts with their more specific counterparts. For instance, a lexical subsumption hierarchy about \emph{Koto} can related the generic classification concept of \emph{Koto} with \emph{kotos} made of different grades of Paulownia wood, each of which is a distinct, more specific classification concept with respect to \emph{Koto}. Last but not the least, the lexical subsumption hierarchy is constructed by modelling \emph{lexical genus-differentiae} \cite{UKC-CICLING} out of linguistically grounded properties of classification concepts encoded in their glosses. To take an example, while the lexical genus of the classification concept \emph{Koto} is \emph{``made of wood"}, the lexical differentia \emph{``type of wood"} helps differentiate (the classification concept of) \emph{kotos} made of Paulownia wood from (the classification concept of) \emph{kotos} made of other woods in the lexical subsumption hierarchy.

%

%



\section{Visual Semantics $\neq$ Lexical Semantics}
\label{S3}
The root of the (in)correspondence between visual and lexical semantics lies in the Semantic Gap Problem (SGP) which has been formally defined as \emph{``the lack of coincidence between the information that one can extract from the visual data} [encoded by visual semantics] \emph{and the interpretation that the same data have for a user} [encoded by lexical semantics] \emph{in a given situation"} \cite{SGP}.

\begin{figure}[htp]
\includegraphics[width=12cm,height=3cm]{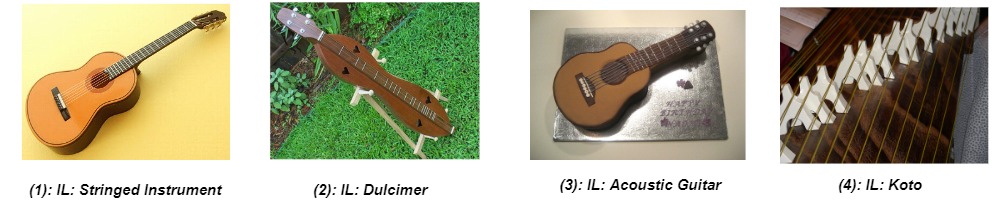}
\caption{Samples from ImageNet Musical Instrument Hierarchy (IL: \textit{ImageNet Label)}}
\centering
\label{I1}
\end{figure}

There are three key dimensions (the combinations of) which result in the lack of coincidence as mentioned above. Firstly, given a target reality (such as an image of a musical instrument), the fact that its visual properties doesn't necessarily coincide with its linguistic properties as defined in a standard gloss. This generates different non-coinciding substance and classification concepts for the same image. Secondly, as a consequence of the above, the visual subsumption hierarchy of substance concepts doesn't necessarily coincide with the lexical subsumption hierarchy of classification concepts formed from the target reality in different languages. Thirdly, the differences in the choice of visual genus-differentiae and of lexical genus-differentiae with respect to the target reality results in potentially many visual subsumption hierarchies, each of which doesn't necessarily coincide with the potentially many lexical subsumption hierarchies. We briefly exemplify below some cases of incorrespondence between visual and lexical semantics (see Fig.\ref{I1}) in the musical instruments hierarchy of ImageNet \cite{IMAGENET-2009} as detailed in \cite{VGTC}.

Let us concentrate on image (1) in Fig.\ref{I1}. Though the image is classified as a good image in ImageNet \cite{ICML-2020} with a one-to-one correspondence between its visual and lexical semantics, it isn't necessarily the case. Depending on the focus of the visual properties of the image, it can be visually classified as different substance concepts such as \emph{stringed instrument} (presence of taut strings), \emph{guitar} (presence of six taut strings) and \emph{acoustic guitar} (absence of output jack). Out of the three cases above, only \emph{stringed instrument} respects a one-to-one alignment between visual and lexical semantics with respect to the ImageNet label. Further, for the same image, there can be different visual differentiae such as \emph{number of strings}, \emph{body shape} or \emph{colour} which can, in turn, generate mutually incompatible visual and lexical subsumption hierarchies. Similarly, image (2) can be visually classified as \emph{stringed instrument} (presence of taut strings), \emph{dulcimer} (presence of four taut strings) and \emph{Apalachian dulcimer} (unique body shape), out of which only \emph{dulcimer} respects the alignment between visual and lexical semantics with respect to the ImageNet label.

The image (3) in Fig.\ref{I1} is an example of mislabelled images in the musical instrument hierarchy of the ImageNet. It is clear from the image that it is of a birthday cake shaped in the form of a guitar (and thus should ideally be encoded as a \emph{birthday cake}) at the visual semantic level. However, we notice a direct incorrespondence between its visual and lexical semantics in the form of a wrong (ImageNet) label titled \emph{Acoustic Guitar}. Image (4), on the other hand, is a rather ambiguous image in the ImageNet. It can be visually classified on the basis of at least two visual property - \emph{number of strings} and \emph{number of movable bridges}. Each of the above properties would, in turn, generate a visual subsumption hierarchy of substance concepts inconsistent with the other and with lexical subsumption hierarchies in different languages.

\section{Aligning Visual \& Lexical Semantics}
\label{S4}
Given the exemplification of the misalignment between visual semantics and lexical semantics in CV systems, we now introduce a \emph{general, domain-agnostic methodology} to enforce alignment between the two aforementioned semantics in such systems. The methodology is organized into four ordered steps:

\begin{enumerate}
    \item \textit{Step (1): Substance Concept Recognition.} Identify the relevant substance concepts in the visual data (such as in an image) by determining their relevant \textit{visual properties}.
    \item \textit{Step (2): Visual Classification.} Organize the recognized substance concepts into a visual subsumption hierarchy by determining visual genus-differentia.
    \item \textit{Step (3): Linguistic classification.} Choose the language labels for each concept in the visual subsumption hierarchy, thus transforming each such substance concept into a classification concept.
    \item \textit{Step (4): Conceptual Classification.} Choose an \emph{alinguistic} identifier to uniquely disambiguate each classification concept denoted by the selected label.
\end{enumerate}
We now elaborate and exemplify each of the above steps one by one.

\textit{Step (1): Substance Concept Recognition.} This step is focused on recognizing substance concepts from substances. Visual perception, which is the reference of a percept to the substance outside the mind \cite{ISKO}, is incrementally facilitated by pure percepts and compound percepts \cite{ISKO}. Given an encounter with a substance from a particular partial view, pure percepts are meaningful impressions (i.e., visual properties) of the substance generated by a single primary sense (vision) and deposited in a memory. Pure percepts, over several sets of encounters, aggregate to form compound percepts. Compound percepts, being agglomerated (meaningful) impressions generated by the association of several pure percepts, are essentially the substance concepts. Such evolving assimilation of newly perceived percepts with pre-existing concepts in the memory is stored and updated in a continuously evolving \emph{cumulative memory} \cite{SNCS-2021}. 

Let us exemplify the above step with case of three stringed musical instruments - \emph{Guitar}, \emph{Flute} and \emph{Koto} (see Fig.\ref{I1} for reference). From a certain distance, all the three can be definitively visually recognized as the substance concept \emph{musical instruments} given their visual property of \emph{means of producing music}. From a nearer distance, though, \emph{Guitar} and \emph{Koto} can be recognized as the substance concept \emph{stringed musical instrument} and \emph{Flute} as the substance concept \emph{wind musical instrument} (the visual property being the different means of producing sound). Finally, from an even closer examination, even the substance concepts \emph{Guitar} and \emph{Koto} can be visually differentiated (the visual property being the number of taut strings). Notice that the substance concept \emph{Guitar} and \emph{Koto} are aggregations of several pure percepts perceived from different partial views deposited in the cumulative memory.

\textit{Step (2): Visual Classification.} The visual classification step, instead, is focused on the modelling of the substance concepts recognized in the previous step into a visual subsumption hierarchy. Such a hierarchy is constructed by organizing substance concepts as a function of their visual properties via exploiting visual genus-differentiae. Let us continue with the previous example of a \emph{Guitar}, \emph{Flute} and \emph{Koto}. The root of the visual subsumption hierarchy will be the substance concept \emph{musical instruments}. On applying the visual differentia \emph{means of sound production} to the visual genus \emph{musical instruments}, the children substance concepts would then be \emph{stringed musical instrument} (the means being \emph{taut strings}) and \emph{wind musical instrument} (the means being \emph{sound pipe}). Further on, the visual genus \emph{stringed musical instrument} can have children substance concepts - \emph{Guitar} (six taut strings) and \emph{Koto} (thirteen taut strings) based on the visual differentia - \emph{the number of taut strings}. Notice that visual genus-differentia in this step is mapped one-to-one with the visual properties considered in the previous step.

\textit{Step (3): Linguistic Classification.} The focus of the linguistic classification step is to exploit language labels, \emph{viz.} words from an appropriate natural language or domain language to annotate the visual subsumption hierarchy as constructed above. This generates what we model as the lexical subsumption hierarchy composed of classification concepts at each level of the hierarchy. For example, the labelling of previous visual subsumption hierarchy of musical instruments in the Italian language will generate its Italian lexical subsumption hierarchy. The labelling of the same visual subsumption hierarchy in Japanese will, instead, generate its Japanese lexical subsumption hierarchy. It is worth noting that the lexical subsumption hierarchies are distinct for each language due to lexical gaps (as studied in \cite{UKC-CICLING,UKC-IJCAI}), but can also be highly \emph{egocentric} depending on one single individual’s purpose \cite{ecai}. Further, the classification concepts and the lexical subsumption hierarchy \textit{definitionally} correspond one-to-one to the substance concepts and visual subsumption hierarchy from the previous step.

\textit{Step (4): Conceptual Classification.} The conceptual classification step formally identifies each concept in the lexical subsumption hierarchy with unique \emph{alinguistic or language independent identifiers}, thus mitigating any remaining linguistic phenomena such as polysemy, homonymy and synonymy \cite{UKC-CICLING}. In the process, this step also facilitates disambiguation of classification concepts from language labels to uniquely machine-identifiable classification concepts for computational reasoning, analytics and communication purposes. For example, the classification concept \emph{Guitar} and \emph{Koto} can be assigned the unique identifers \texttt{1278956} and \texttt{9998705} respectively to computationally disambiguate them.

There are three crucial observations. Firstly, the fact that the four steps (individually and in unison) of the methodology enforce a one-to-one alignment between visual and lexical semantics, thus eliminating the many-to-many mapping between them existing in a particular context. The experiment design and execution reported in \cite{VGTC} for (a subset of) the musical instruments hierarchy of the ImageNet elucidates how such an alignment is enforced in practice. Secondly, notice that the algorithmic implementation of the methodology (see \cite{SNCS-2021,VGTC} for implementation and preliminary results) as described above is \emph{continual} and \emph{domain-agnostic} in nature. A CV system grounded in the methodology would thus be able to, for any domain, continually recognize substance concepts and incrementally generate, refine and disambiguate substance and classification concept hierarchies. Recent efforts in the direction include the ongoing project \emph{Media UKC} \cite{ISKO}, a multimedia extension of the multilingual lexical semantic resource \emph{Universal Knowledge Core (UKC)} \cite{UKC-CICLING,UKC-IJCAI}. Finally, as reported in the experiments performed in \cite{VGTC}, there is a \emph{considerable improvement} in (image) classification accuracy when state-of-the-art (CV) neural network models are trained on \emph{ground truth} generated following the proposed methodology above.

\section{Conclusion}
\label{S5}
The paper introduced \emph{Visual Semantics} and \emph{Lexical Semantics} in the context of CV systems and exhaustively exemplified the misalignment between them. As a solution to such misalignment, the paper introduced a general, domain-agnostic methodology over four ordered steps to enforce alignment between the two kinds of semantics in a particular context.

\section*{Acknowledgement}
The research conducted by Fausto Giunchiglia, Mayukh Bagchi and Xiaolei Diao has received funding from the \emph{“DELPhi - DiscovEring Life Patterns”} project supported by the MIUR (PRIN) 2017.

\bibliographystyle{splncs04}
\bibliography{iConference}

\end{document}